\documentclass[11pt,a4paper]{article}
\usepackage{pdfpages}
\usepackage[hyperref]{emnlp-ijcnlp-2019}
\usepackage{times}
\usepackage{ifthen}
\usepackage{latexsym}
\usepackage{amsmath}
\usepackage{multirow}
\usepackage{booktabs}
\usepackage{arydshln}
\usepackage{subcaption}
\usepackage{url}

\aclfinalcopy 

\title{Don't Take the Easy Way Out:\\ 
Ensemble Based Methods for Avoiding Known Dataset Biases}

\author{Christopher Clark$^{*}$, Mark Yatskar$^\dagger$, Luke Zettlemoyer$^{*}$ \\ \\
$^*$Paul G. Allen School of CSE, University of Washington \\
{\tt \{csquared, lsz\}@cs.uw.edu} \\ \\
$^\dagger$Allen Institute for Artificial Intelligence, Seattle WA \\
{\tt marky@allenai.org} \\
}

\date{}

\begin{document}

\maketitle

\newcommand{\tablefont}{\small}

\begin{abstract}
State-of-the-art models often make use of superficial patterns in the data that do not generalize well to out-of-domain or adversarial settings.
For example, textual entailment models often learn that particular key words imply entailment, irrespective of context, and visual question answering models learn to predict prototypical answers, without considering evidence in the image.
In this paper, we show that if we have prior knowledge of such biases, we can train a model to be more robust to domain shift.
Our method has two stages: we (1) train a naive model that makes predictions exclusively based on dataset biases, and (2) train a robust model as part of an ensemble with the naive one in order to encourage it to focus on other patterns in the data that are more likely to generalize.
Experiments on five datasets with out-of-domain test sets show significantly improved robustness in all settings, including a 12 point gain on a changing priors visual question answering dataset and a 9 point gain on an adversarial question answering test set.
\end{abstract}

\section{Introduction}
\begin{figure*}
    \centering
    \includegraphics[width=.92\textwidth]{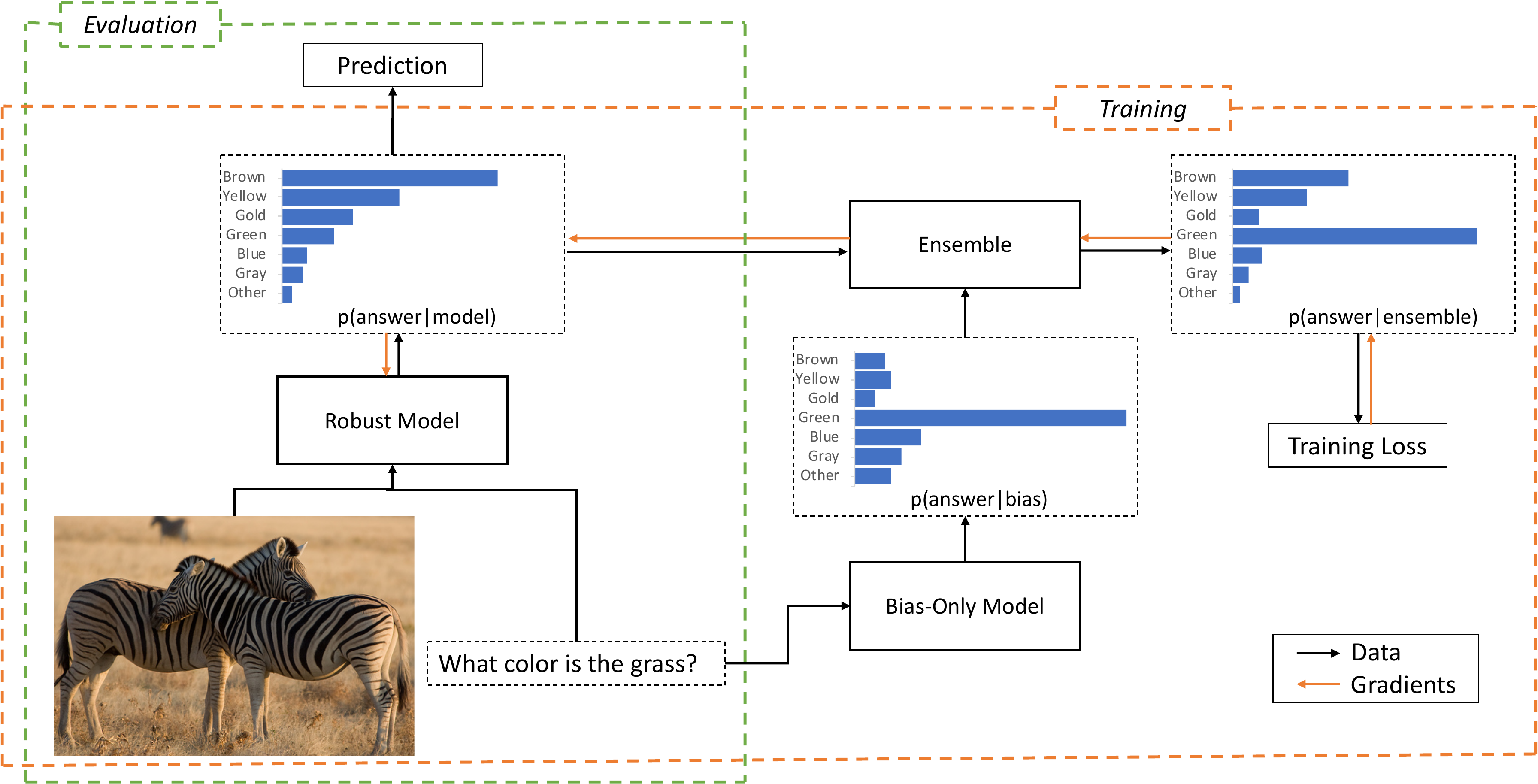}
    \caption{An example of applying our method to a Visual Question Answering (VQA) task. We assume predicting green for the given question is almost always correct on the training data. 
    To prevent a model from learning this bias, we first train a bias-only model that only uses the question as input, and then train a robust model in an ensemble with the bias-only model.
    Since the bias-only model will have already captured the target pattern, the robust model has no incentive to learn it, and thus does better on test data where the pattern is not reliable.}
    \label{fig:summary}
\end{figure*}

While recent neural models have shown remarkable results, these achievements have been tempered by the observation that they are often exploiting dataset-specific patterns that do not generalize well to out-of-domain or adversarial settings.
For example, entailment models trained on MNLI~\cite{bowman2015large} will guess an answer based solely on the presence of particular keywords~\cite{gururangan2018annotation} or whether sentences pairs contain the same words~\cite{mccoy2019right}, while QA models trained on SQuAD~\cite{squad} tend to select text near question-words as answers, regardless of context~\cite{adversarial_squad}.

We refer to these kinds of superficial patterns as  bias.
Models that rely on bias can perform well on in-domain data, but are brittle and easy to fool (e.g., SQuAD models are easily distracted by irrelevant sentences that contain many question words). 
Recent concern about dataset bias has led researchers to re-examine many popular datasets, resulting in the discovery of a wide variety of biases~\cite{vqa_cp,anand2018blindfold,min2019compositional,schwartz2017effect}.  

In this paper, we build on these works by showing that, once a dataset bias has been identified, we can improve the out-of-domain performance of models by preventing them from making use of that bias.
To do this, we use the fact that these biases can often be explicitly modelled with simple, constrained baseline methods to factor them out of a final model through ensemble-based training.


Our method has two stages. First, we build a bias-only model designed to capture a naive solution that performs well on the training data, but generalizes poorly to out-of-domain settings. 
Next, we train a second model in an ensemble with the pre-trained bias-only model, which incentivizes the second model to learn an alternative strategy, and use the second model alone on the test set.
We explore several different ensembling methods, building on product-of-expert style approaches~\cite{hinton1999products,smith2005logarithmic}.
Figure~\ref{fig:summary} shows an example of applying this procedure to prevent a visual question answering (VQA) model from guessing answers because they are typical for the question, a flaw observed in 
VQA models~\cite{vqa2,vqa_cp}.

We evaluate our approach on a diverse set of tasks, all of which require models to overcome a challenging domain-shift between the train and test data.
First, we build a set of synthetic datasets that contain manually constructed biases by adding artificial features to MNLI. 
We then consider three challenge datasets proposed by prior work~\cite{vqa_cp,mccoy2019right,adversarial_squad}, which were designed to break models that adopt superficial strategies on well known textual entailment~\cite{bowman2015large}, reading comprehension~\cite{squad}, and VQA~\cite{vqa1} datasets.

We additionally construct a new QA challenge dataset, TriviaQA-CP (for TriviaQA changing priors). This dataset was built by holding out questions from TriviaQA~\cite{joshi2017triviaqa} that ask about particular kinds of entities from the train set, and evaluating on those questions in the dev set, in order to challenge models to generalize between different types of questions.
 
We are able to improve out-of-domain performance in all settings, including a 6 and 9 point gain on the two QA datasets. 
On the VQA challenge set, we achieve a 12 point gain, compared to a 3 point gain from prior work.
In general, we find using an ensembling method that can dynamically choose when to trust the bias-only model is the most effective, and we present synthetic experiments and qualitative analysis to illustrate the advantages of that approach.
We release our datasets and code to facilitate future work.\footnote{github.com/chrisc36/debias} 

\section{Related Work}
Researchers have raised concerns about bias in many datasets. 
For example, many joint natural language processing and vision datasets can be partially solved by models that ignore the vision aspect of the task~\cite{jabri2016revisiting, zhang2016yin,anand2018blindfold,caglayan2019probing}.
Some questions in recent multi-hop QA datasets~\cite{hotpotqa,welbl2018constructing} can be solved by single-hop models~\cite{chen2019understanding,min2019compositional}.
Additional examples include story completion~\cite{schwartz2017effect} and multiple choice questions~\cite{clark2016combining,clark2018think}.
Recognizing that bias is a concern in diverse domains, our work is the first to perform an evaluation across multiple datasets spanning language and vision.

Recent dataset construction protocols have tried to avoid certain kinds of bias. For example, both CoQA~\cite{reddy2018coqa} and QuAC~\cite{choi2018quac} take steps to prevent annotators from using words that occur in the context passage, VQA 2.0~\cite{vqa2} selects examples to limit the effectiveness of question-only models, and others have filtered examples solvable by simple baselines~\cite{hotpotqa,record,clark2018think,swag}.
While reducing bias is important, developing ways to prevent models from using known biases will allow us to continue to leverage existing datasets, and update our methods as our understanding of what biases we want to avoid evolve.

Recent work has focused on biases that come from ignoring parts of the input (e.g., guessing the answer to a question before seeing the evidence).
Solutions include generative objectives to force models to understand all the input~\cite{generative_qa}, carefully designed model architecture~\cite{vqa_cp,zhang2016yin}, or adversarial removal of class-indicative features from model's internal representations~\cite{ramakrishnan2018overcoming,zhang2018mitigating,belinkov2019adversarial,grand2019adversarial}.
In contrast, we consider biases beyond partial-input cases~\cite{feng2019misleading}, and show our method is superior on VQA-CP.
Concurrently, \citet{HeHe} also suggested using a product-of-experts ensemble to train unbiased models, but we consider a wider variety of ensembling approaches and test on additional domains.

A related task is preventing models from using particular problematic dataset features, which is often studied from the perspective of fairness~\cite{zhao2017men, burns2018women}. 
A popular approach is to use an adversary to remove information about a target feature, often gender or ethnicity, from a model's internal representations~\cite{edwards2015censoring, wang2018adversarial, kim2018learning}.
In contrast, the biases we consider 
are related to features that are essential to the overall task, so they cannot simply be ignored.

Evaluating models on out-of-domain examples built by applying minor perturbations to existing examples has also been the subject of recent study~\cite{szegedy2013intriguing,belinkov2017synthetic,carlini2018audio,glockner2018breaking}. The domain shifts we consider involve larger changes to the input distribution, built to uncover higher-level flaws in existing models. 


\section{Methods}
\newcommand{\vecsym}[1]{#1}

\label{sect:methods}
This section describes the two stages of our method, (1) building a bias-only model and (2) using it to train a robust model through ensembling. 

\subsection{Training a Bias-Only Model}
The goal of the first stage is to build a model that performs well on training data, but is likely to perform very poorly
on the out-of-domain test set. 
Since we assume we do not have access to examples from the test set, we must apply \textit{a-priori} knowledge to meet this goal.

The most straightforward approach is to identify a set of features that are correlated with the class label during training, but are known to be uncorrelated or anticorrelated with the label on the test set, and then train a classifier on those features.\footnote{Since the bias-only model is trained on the same train-set as the robust model care should also be taken to minimize overfitting, although the bias-only model is typically simple enough that this is not an issue.} For example, our VQA-CP~\cite{vqa_cp} bias-only model (see Section~\ref{sect:vqap_cp}) uses the question type as input, because the correlations between question types and answers is very different in the train set than the test set
(e.g., 2 is a common answer to ``How many..." questions on the train set, but is rare for such questions on the test set).

However, a benefit of our method is that the bias can be modelled using any kind of predictor, giving us a way to capture more complex intuitions. 
For example, on SQuAD our bias-only model operates on a view of the input built from TF-IDF scores (see Section~\ref{sect:adverserial_squad}), and on our changing prior TriviaQA dataset our bias-only model makes use of a pre-trained named entity recognition (NER) tagger (see Section~\ref{sect:triviaqa_cp}).

\subsection{Training a Robust Model}
This stage trains a robust model that avoids using the method learned by the bias-only model.

\subsubsection{Problem Definition} We assume $n$ training examples $\langle x_1, x_2, \ldots, x_n\rangle$, each of which has an integer label $y_i$, where $y_i \in \{1, 2, \ldots, C\}$ and $C$ is the number of classes. 
We additionally assume a pre-trained bias-only predictor, $h$, where $h(x_i) = \vecsym{\vecsym{b_i}} = \langle b_{i1}, b_{i2}, .. b_{iC} \rangle$ and $b_{ij}$ is the bias-only model's predicted probability of class $j$ for example $i$.
Finally we have a second predictor function, $f$, with parameters $\theta$, where $f(x_i, \theta) = \vecsym{p_i}$ and $\vecsym{p_i}$ is a similar probability distribution over the classes.
Our goal is to construct a training objective to optimize $\theta$ so that $f$ will learn to select the correct class without using the strategy captured by the bias-only model. 

\subsubsection{General Approach} 
We train an ensemble of $h$ and $f$. 
In particular, for each example, a new class distribution, $\vecsym{\hat{p_i}}$, is computed by combining $\vecsym{p_i}$ and $\vecsym{b_i}$. During training, the loss is computed using $\vecsym{\hat{p_i}}$ and the gradients are backproped through $f$. During evaluation $f$ is used alone. We propose several different ensembling methods.

\subsubsection{Bias Product} 
 Our simplest ensemble is a product of experts~\cite{hinton1999products}:
 $$\vecsym{\hat{p_i}} = \mathit{softmax}(\log(\vecsym{p_i}) + \log(\vecsym{b_i}))$$
 
 Equivalently, $\vecsym{\hat{p_i}} \propto \vecsym{p_i} \circ \vecsym{b_i}$, where $\circ$ is elementwise multiplication.
\\
\\ \noindent
\textbf{Probabilistic Justification}:
For a given example, $x$, let $x^b$ be the bias of the example. That is, it is the features we will use in our bias-only model. Let $x^{-b}$ be a view of the example that captures all information about that example except the bias. Assume that $x^{-b}$ and $x^b$ are conditionally independent given the label, $c$. Then to compute $p(c|x)$ we have:
\begin{eqnarray}
p(c|x) & = & p(c|x^b, x^{-b})  \\
&\propto& p(c|x^{-b}) p(x^b|c,x^{-b})  \\
&= & p(c|x^{-b}) p(x^b|c) \\
&= &p(c|x^{-b})\dfrac{p(c|x^b)p(x^b)}{p(c)} \\
&\propto&p(c|x^{-b})\dfrac{p(c|x^b)}{p(c)}
\end{eqnarray}

Where 2 is from applying Bayes Rule while conditioning on $x^{-b}$, 3 follows from the conditional independence assumption, and 4 applies Bayes Rule a second time to $p(x^b|c)$.

We cannot directly model $p(c|x^{-b})$ because it is usually not possible to create a view of the data that excludes the bias. 
Instead, with the goal of encouraging the model to fall into the role of computing $p(c|x^{-b})$, we compute ${p(c|x^b)}/{p(c)}$ using the bias-only model, and train the product of the two models to compute $p(c|x)$. 

In practice, we ignore the $p(c)$ factor because, on our datasets, either the classes are uniformly distributed (MNLI), the bias-only model cannot easily capture a class prior since it is using a pointer network (QA), or because we want to remove class priors from model anyway (VQA).

\subsubsection{Learned-Mixin}
The assumption of conditional independence (Equation 3) will often be too strong.
For example, in some cases the robust model might be able to predict the bias-only model will be unreliable for certain kinds of training examples.
We find that this can cause the robust model to selectively adjust its behavior in order to compensate for the inaccuracy of the bias-only model, leading to errors in the out-of-domain setting (see Section~\ref{sect:synthetic}).

Instead we allow the model to explicitly determine how much to trust the bias given the input: 
$$\vecsym{\hat{p_i}} = \mathit{softmax}(\log(\vecsym{p_i}) + g(x_i)\log(\vecsym{b_i}))$$ 
where $g$ is a learned function. We compute $g$ as $\mathit{softplus}(w\cdot h_i)$ where $w$ is a learned vector, $h_i$ is the last hidden layer of the model for example $x_i$, and the $\mathit{softplus}(x) = \log(1 + e^x)$ function is used to prevent the model reversing the bias by multiplying it by a negative weight. $w$ is trained with the rest of the model parameters. This reduces to bias product when $g(x_i) = 1$.

A difficulty with this method is that the model could learn to integrate the bias into $\vecsym{p_i}$ and set $g(x_i) = 0$.
We find this does sometimes occurs in practice, and our next method alleviates this challenge.

\subsubsection{Learned-Mixin +H}
To prevent the learned-mixin ensemble from ignoring $b_i$, we add an entropy penalty to the loss:

$$\mathit{R} = wH(\mathit{softmax}(g(x_i)\log(\vecsym{b_i})))$$ 

Where $H(z) = -\sum_{j}z_{j}\log(z_{j})$ is the entropy and $w$ is a hyperparameter. Penalizing the entropy encourages the bias component to be non-uniform, and thus have a greater impact on the ensemble.

\section{Evaluation Methodology}
\begin{table*}[ht]
    \newcommand{\summarysp}{\addlinespace[0.12cm]}
    \centering
     \begin{small}
    \begin{tabular}{llp{4cm}llll}  \toprule
         Task & Dataset & Domain Shift & Bias-Only Model & Main Model \\ \midrule
         NLI & Synthetic MNLI & Synthetic indicator features are randomized & Indicator features & Co-Attention \\ \summarysp
         VQA & VQA-CP v2.0 & Correlations between question-types and answers are altered & Question-type & BottomUpTopDown \\ \summarysp
         NLI & HANS & Sentence pairs always contain the same words & Shared word features & BERT \& Co-Attention \\ \summarysp
         QA & Adv. SQuAD & Distractor sentences are added to the context & TF-IDF sentence selector & Modified BiDAF \\ \summarysp
         QA & TriviaQA-CP & Questions ask about different kinds of entities & NER answer detector & Modified BiDAF \\ 
         \bottomrule
    \end{tabular}
    \end{small}
    \caption{Summary of the evaluations we perform, Domain Shift refers to what changes between the train and test data, and Bias-Only Model specifies how the bias model we use was constructed. See the main text for details.}
    \label{tab:evaluations}
\end{table*}

We evaluate our methods on several datasets that have out-of-domain test sets.
Some of these tasks, such as HANS~\cite{mccoy2019right} or Adversarial SQuAD~\cite{adversarial_squad}, can be solved easily by generating additional training examples similar to the ones in the test set (e.g., \citet{wang2018robust}).
We, instead, demonstrate that it is possible to improve performance on these tasks by exploiting knowledge of general, biased strategies the model is likely to adopt.

Our evaluation setup consists of a training set, an out-of-domain test set, a bias-only model, and a main model. To run an evaluation we train the bias-only model on the train set, train the main model on the train set while employing one of the methods in Section~\ref{sect:methods}, and evaluate the main model on the out-of-domain test set. We also report performance on the in-domain test set, when available. 
We use models that are known to work well for their respective tasks for the main model, and do not further tune their hyperparameters or perform early stopping.

We consider two extractive QA datasets, which we treat as a joint classification task where the model must select the start and end answer token~\cite{wang2016machine}. 
For these datasets, we build independent bias-only models for selecting the start and end token, and separately ensemble those biases with the classifier's start token and end token output distributions.  
We apply a ReLU layer to the question and passage embeddings, followed by max-pooling, to construct a hidden state for computing the learned-mixin weights.

We compare our methods to a reweighting baseline described below, and to training the main model without any modifications. On VQA we also compare to the adversarial methods from~\citet{ramakrishnan2018overcoming} and ~\citet{grand2019adversarial}. The other biases we consider are not based on observing only part of the input, so these adversarial methods cannot be directly applied.

\subsection{Reweight Baseline}
As a non-ensemble baseline, we train the main model on a weighted version of the data, where the weight of example $x_i$ is $1 - b_{iy_i}$ (i.e., we weigh examples by one minus the probability the bias-only model assigns the correct label). 
This encourages the main model to focus on examples the bias-only model gets wrong.
    
\subsection{Hyperparameters}
One of our methods (Learned-Mixin +H) requires hyperparameter tuning. However hyperparameter tuning is challenging in our setup since our assumption is that we have no access to out-of-domain test examples during training. 
A plausible option would be to tune hyperparameters on a dev set that exhibits a related, but not identical, domain shift to the test set, but unfortunately none of our datasets have such dev sets. Instead we follow prior work~\cite{grand2019adversarial,ramakrishnan2018overcoming} and perform model selection on the test set. Although this presents an important caveat to the results of this method, we think it is still of interest to observe that the entropy regularizer can be very impactful. Future work may be able to either construct suitable development sets, or propose other hyperparameter-tuning methods to relieve this issue. The hyperparameters selected are shown in Appendix~\ref{appendix:penalty_weights}.


%

\section{Experiments}
\newcommand{\resultheader}[1]{~\\\noindent\textbf{#1}:}
\newcommand{\resultheaderns}[1]{\noindent\textbf{#1}:}
We provide experiments on five different domains, summarized in Table~\ref{tab:evaluations}, each of which requires models to overcome a challenging domain-shift between train and test data.
In the following sections we provide summaries of the datasets, main models and bias-only models, but leave low-level details to the appendix. 

\subsection{Synthetic Data}
\label{sect:synthetic}
\resultheaderns{Data} We experiment with a synthetic dataset built by modifying  MNLI~\cite{bowman2015large}. In particular, we add a feature that is correlated with the class label to the train set, and build an out-of-domain test set by adding a randomized version of that feature to the MNLI matched dev set. We additionally construct an in-domain test set by modifying the matched dev set in the same way as was done in the train set. We build three variations of this dataset:
\\
\\  
\textit{Indicator}: Adds the token ``0", ``1", or ``2" to the start of the hypothesis, such that 80\% of the time the token corresponds to the example's label (i.e., ``0" if the class is ``entailment", ``1" if the class is contradiction, ect.). In the out-of-domain test set, the token is selected randomly.
\\
\\
\textit{Excluder}: The same as Indicator, but with a 3\% chance the added token corresponds to the example's label, meaning the token can usually be used to eliminate one of the three output classes.
\\
\\
\textit{Dependent}: In the previous two settings, the added bias is independent of the example given the example's label. To simulate a case where this independence is broken, we experiment with adding an additional feature that is correlated with the bias feature, but is not treated as being part of the bias (i.e., it is not used by the bias-only model).
In particular, 80\% of the time a token is added to the start of the hypothesis that matches the label with 90\% probability, and the ``0" token is appended to the end of the hypothesis. The other 20\% of the time a random token is prepended and ``1" is appended. 

\resultheader{Bias-Only Model} The bias-only model predicts the label using the first token of the hypothesis. 

\resultheader{Main Model} We use a recurrent co-attention model, similar to ESIM~\cite{chen2016enhanced}. Details are given in Appendix~\ref{appendix:nli_model}.

\begin{table*}
\centering
\tablefont
\begin{tabular}{lcccccc} \toprule
\multirow{2}{*}{Debiasing Method} & \multicolumn{2}{c}{Indicator} & \multicolumn{2}{c}{Excluder} & \multicolumn{2}{c}{Dependent}\\ 
 & Acc. & w/Bias & Acc. & w/Bias & Acc. & w/Bias\\ \midrule
None & 69.36 & 86.49 & 68.06 & 83.56 & 63.23 & 87.90 \\
Reweight & 75.44 & 82.74 & 70.36 & 83.29 & 69.81 & 85.50 \\
Bias Product & 76.27 & 81.32 & 77.33 & 80.41 & 71.85 & 84.98 \\
Learned-Mixin & 76.29 & 81.35 & 77.80 & 78.86 & 75.75 & 77.70 \\
Learned-Mixin +H & 76.77 & 77.65 & 77.90 & 78.57 & 75.79 & 76.65 \\ \hdashline
Unbiased Training & 78.94 & 78.94 & 78.94 & 78.94 & 78.86 & 78.86 \\
\bottomrule
\end{tabular}

\caption{Results on MNLI with different kinds of synthetic bias. The Acc columns show the accuracy on the out-of-domain test set, and the w/Bias columns show accuracy on the in-domain test. Unbiased Training is an upper bound constructed by training a model with the same randomized features that are used at test time.}
\label{tab:synthetic}
\end{table*}
 
\resultheader{Results} Table~\ref{tab:synthetic} shows the results.
All ensembling methods work well on the Indicator bias. The reweight method performs poorly on the Excluder bias, likely because the bias-only model assigns the correct class approximately 50\% probability for almost all the training examples, making the weights mostly uniform. 
This illustrates a general weakness with reweighting methods: they require at least a small number of bias-free examples for the model to learn from.

The bias product method performs poorly on the Dependent bias.
Inspection shows that, when the indicator is 1, the bias product model is anti-correlated with the bias. In particular, it assigns an average of $22.5\%$ probability to the class indicated by the bias, where an unbaised model would assign an average of $33\%$ since the bias is random.
The root cause is that, if the indicator is 1, the model knows the bias is likely to be wrong, so it learns to subtract the value the bias-only model will produce from its own output in order to cancel out the bias-only model's effect on the ensemble's output. 

The learned-mixin model does not suffer from this issue, and assigns the class indicated by the bias an average of $34.5\%$ probability. Analysis shows that $g(x_i)$ is set to $0.00 \pm 0.0001$ when the indicator is turned off, and to $1.91 \pm 0.285$ otherwise, showing that the model learns to turn off the bias-only component of the ensemble as needed, thus avoiding this over-compensating issue. The entropy regularizer appears to be unnecessary on this dataset because $g(x_i)$ does not go to zero.

\subsection{VQA-CP}
\label{sect:vqap_cp}

\begin{table}
\centering
\tablefont
\begin{tabular}{lc} \toprule
Debiasing Method & Acc. \\ \midrule
None & 39.18\\
Reweight & 40.06\\
Bias Product & 39.93\\
Learned-Mixin & 48.69\\
Learned-Mixin +H & 52.05\\ \hdashline
\citet{ramakrishnan2018overcoming} & 41.17\\
\citet{grand2019adversarial} & 42.33\\
\bottomrule
\end{tabular}
\caption{Results on the VQA-CP v2.0 test set.}
\label{tab:vqa_results}
\end{table}

\resultheaderns{Data} We evaluate on the VQA-CP v2~\cite{vqa_cp} dataset, which was constructed by re-splitting the VQA 2.0~\cite{vqa2} train and validation sets into new train and test sets such that the correlations between question types and answers differs between each split. For example, ``tennis" is the most common answer for questions that start with ``What sport..." in the train set, whereas ``skiing" is the most common answers for those questions in the test set. Models that choose answers because they are typical in the training data will perform poorly on this test set.

\resultheader{Bias-Only Model} VQA-CP comes with questions annotated with one of 65 question types, corresponding to the first few words of the question (e.g., ``What color is''). The bias-only model uses this categorical label as input, and is trained on the same multi-label objective as the main model.

\begin{figure*}[t]
    \centering
    \includegraphics[width=\textwidth]{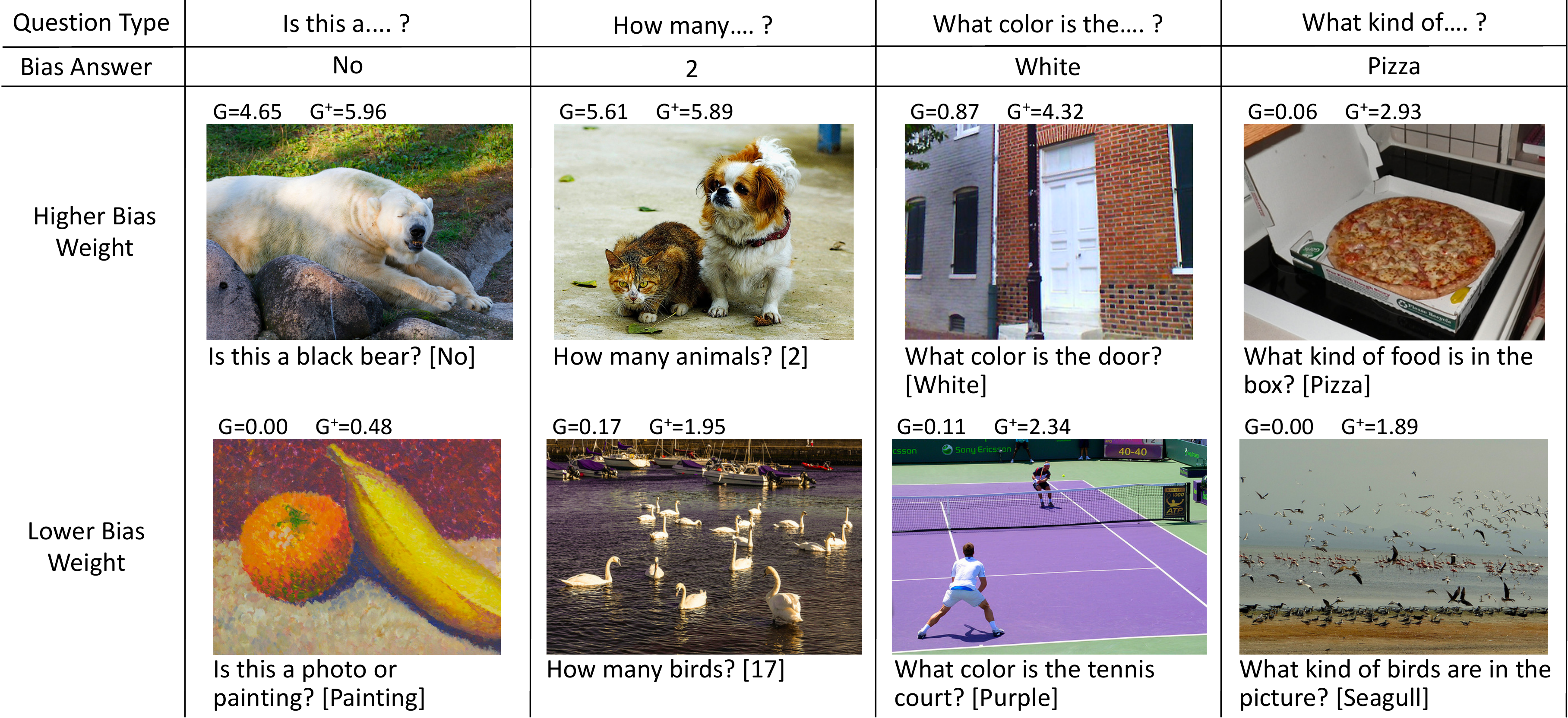} 
    \caption{Qualitative examples of the values of $g(x_i)$ on the VQA-CP training data for the learned-mixin model (labelled ``G'') and learned-mixin +H model (labelled ``G+''). The question type and the bias model's highest ranked answer for that type are shown above. We find $g(x_i)$ is larger when the bias answers are likely to be correct.}
    \label{fig:vqa_qualitative}
\end{figure*}

\resultheader{Main Model} We use a popular implementation\footnote{github.com/hengyuan-hu/bottom-up-attention-vqa} of the BottomUpToDown~\cite{bottomuptopdown} VQA model. This model uses a multi-label objective, so we apply our ensemble methods by treating each possible answer as a two-class classification problem.\footnote{Since the bias sometimes assigns a zero probability to an answer, we additionally add $\sigma(\alpha)$ to the bias probabilities where $\alpha$ is learned parameter to allow the model to soften the bias as needed}
\\
\\
\textbf{Results:} Table~\ref{tab:vqa_results} shows the results. The learned-mixin method was highly effective, boosting performance on VQA-CP by about 9 points, and the entropy regularizer can increase this by another 3 points, significantly surpassing prior work. 
For the learned-mixin ensemble, we find $g(x_i)$ is strongly correlated with the bias's expected accuracy\footnote{Computed as $\sum_j s_{ij} b_{ij} / \sum_j b_{ij}$ where $s_{ij}$ is the score for class $j$ on example $i$}, with a spearmanr correlation of 0.77 on the test data. Qualitative examples (Figure~\ref{fig:vqa_qualitative}) further suggest the model increases $g(x_i)$ when it knows if can rely on the bias-only model.

\subsection{HANS}
\resultheaderns{Data} We evaluate on the HANS adversarial MNLI dataset~\cite{mccoy2019right}. This dataset was built by constructing templated examples of entailment and non-entailment, such that the hypothesis sentence only includes words that are also in the premise sentence. Naively trained models tend to classify all such examples as ``entailment" because detecting the presence of many shared words is an effective tactic on MNLI.

\resultheader{Bias-Only Model} The bias-only model is a shallow linear classifier using the following features: (1) whether the hypothesis is a sub-sequence of the premise, (2) whether all words in the hypothesis appear in the premise, (3) the percent of words from the hypothesis that appear in the premise, (4) the average of the minimum distance between each premise word with each hypothesis word, measured using cosine distance with the fasttext~\cite{fasttext_word_vectors} word vectors, and (5) the max of those same distances. We constrain the bias-only model to put the same amount of probability mass on the neutral and contradiction classes so it focuses on distinguishing entailment and non-entailment, and reweight the dataset so that the entailment and non-entailment examples have an equal total weight to prevent a class prior from being learned.

\resultheader{Main Models} We experiment with both the uncased BERT base model~\cite{devlin2018bert}, and the same recurrent model used for the synthetic data (see Appendix~\ref{appendix:nli_model}). We use the default hyper-parameters for BERT since they work well for MNLI.

\begin{table}
    \centering
    \tablefont
\begin{tabular}{lcccc} \toprule
\multirow{2}{*}{Debiasing Method} & \multicolumn{2}{c}{Co-Attention} & \multicolumn{2}{c}{BERT}\\
 & HANS & MNLI & HANS & MNLI\\ \midrule
None & 50.58 & 78.73 & 62.40 & 84.24 \\
Reweight & 52.85 & 77.03 & 69.19 & 83.54 \\
Bias Product & 53.69 & 76.63 & 67.92 & 82.97 \\
Learned-Mixin & 51.65 & 78.05 & 64.00 & 84.29 \\
Learned-Mixin +H & 53.35 & 74.50 & 66.15 & 83.97 \\
\bottomrule
\end{tabular}   
\caption{Accuracy on the adversarial MNLI dataset, HANS, and the MNLI matched dev set.}
\label{tab:hans}
\end{table}
 
\resultheader{Results}
Table~\ref{tab:hans} shows the results. We show scores for individual heuristics used in HANS in Appendix~\ref{appendix:fine_grained_hans}. For the recurrent method, both the bias product and learned-mixin +H methods result in about a three point gain. However, for the BERT model, the simpler reweight method is more effective. 
We noticed high variance in performance between runs in this setting, and speculate the ensemble methods might be compounding this instability by introducing additional complexity.

\subsection{Adversarial SQuAD}
\label{sect:adverserial_squad}
\resultheaderns{Data} We evaluate on the Adversarial SQuAD~\cite{adversarial_squad} dataset, which was built by adding distractor sentences to the passages in SQuAD~\cite{squad}. The sentences are built to closely resemble the question and contain a plausible answer candidate, but with a few key semantic changes to ensure they do not incidentally answer the question. Models that naively focus on sentences that contain many question words are often fooled by the new sentence.

\resultheader{Bias-Only Models} We consider two bias-only models: (1) TF-IDF: the TF-IDF score between each sentence and question is used to select an answer (meaning tokens within the same sentence all get the same score) and (2) TF-IDF Filtered: the same but excluding pronouns and numbers from the words used to compute the TF-IDF scores. The second model is motivated by the fact distractor sentences never include numbers or pronouns that occur in the question.

\resultheader{Main Model} We use an updated version of BiDAF~\cite{seo2016bidirectional}, that uses the fasttext words vectors~\cite{fasttext_word_vectors}, includes an additional recurrent layer, and simplifies the prediction stage (see Appendix~\ref{appendix:bidaf-model}).

\begin{table*}[]
    \centering
    \tablefont
\begin{tabular}{lcccccc} \toprule
\multirow{2}{*}{Debiasing Method} & \multicolumn{3}{c}{TF-IDF Filtered} & \multicolumn{3}{c}{TF-IDF}\\
\cmidrule(r){2-4} \cmidrule(l){5-7}
 & AddSent & AddSentOne & Dev & AddSent & AddSentOne & Dev\\ \midrule
None & 42.54 & 53.91 & 80.61 & 42.54 & 53.91 & 80.61 \\
Reweight & 41.55 & 53.06 & 80.59 & 42.74 & 53.83 & 80.51 \\
Bias Product & 47.17 & 57.74 & 78.63 & 44.41 & 55.73 & 78.22 \\
Learned-Mixin & 42.25 & 53.51 & 80.39 & 42.00 & 53.46 & 80.46 \\
Learned-Mixin +H & 51.84 & 60.66 & 75.94 & 48.30 & 58.26 & 74.14 \\
\bottomrule
\end{tabular}
    \caption{F1 scores on Adversarial SQuAD and the standard SQuAD dev set using two different bias-only models.}
\label{fig:adversarial_squad}
\end{table*}

\resultheader{Results} Table~\ref{fig:adversarial_squad} shows the results. We find the bias product method improves performance by up to 3 points, and the learned-mixin +H model achieves up to a 9 point gain. The importance of including the entropy penalty is explained by the fact that, without the penalty, the model learns to ignore the bias by settings $g(x_i)$ close to zero. 
For example, on the AddSent dataset with the TF-IDF filtered bias, the learned-mixin ensemble sets $g(x_i)$ to an average of $0.13$, while the learned-mixin +H ensemble increases that to $5.16$.
The high values are likely caused by the fact the bias-only model is very weak, since it assigns the same score to each token in each sentence, so the model can often scale it by large values.
As expected, we get better results using the TF-IDF Filtered bias which is more closely tailored to how the test set was constructed.

\subsection{TriviaQA-CP}
\label{sect:triviaqa_cp}
\begin{table}[]
    \centering
    \tablefont
\begin{tabular}{lcccc} \toprule
\multirow{2}{*}{Debiasing Method} & \multicolumn{2}{c}{Location} & \multicolumn{2}{c}{Person}\\
 & CP & Dev & CP & Dev\\ \midrule
None & 41.23 & 59.27 & 39.69 & 55.26 \\
Reweight & 40.14 & 59.18 & 39.96 & 55.38 \\
Bias Product & 44.42 & 60.02 & 40.58 & 55.20 \\
Learned-Mixin & 41.15 & 61.64 & 41.31 & 56.08 \\
Learned-Mixin +H & 47.77 & 57.74 & 44.37 & 54.83 \\
\bottomrule
\end{tabular}
    \caption{EM scores on two changing priors TriviaQA datasets. The CP column shows scores on the changing priors test set, and Dev shows in-domain scores.}
    \label{tab:triviaqa_cp}
\end{table}
\resultheaderns{Data} We construct a changing-prior QA dataset from TriviaQA~\cite{joshi2017triviaqa} by categorizing questions into three classes, Person, Location, and Other, based on what kind of entity they are asking about.
During training, we hold out all the person questions or all the location questions from the train set, and evaluate on the person or location questions in the TriviaQA dev set. Details can be found in Appendix~\ref{appendix:triviaqa_cp}. 

\resultheader{Bias-Only Model} The bias-only model uses NER tags, identified by running the Stanford NER Tagger~\cite{corenlp_ner} on the passage, as input. We only apply the model to tokens that have a NER tag, and assign all other tokens the average score given to the tokens with NER tags in order to prevent the model from reflecting a preference for entity tokens in general.

\resultheader{Main Model} 
We use a larger version of the model used for Adversarial SQuAD (see Appendix~\ref{appendix:bidaf-model}), to account for the larger dataset.

\resultheader{Results} Table~\ref{tab:triviaqa_cp} shows the results. Similar to adversarial SQuAD, the bias product method is moderately effective, and the ensemble method is superior as long as a suitable regularizer is applied. We again observe that the learned-mixin method tends to push $g(x_i)$ close to zero without the entropy penalty (average of $0.25$ without the penalty vs. $5.01$ with the penalty on the Location dev set). We see smaller gains on the person dataset. One possible cause is that differentiating between people and other named entities, such as organizations or groups, is difficult for the main model, and as a result it does not learn a strong non-person prior even without the use of a debiasing method.

\subsection{Discussion}
Despite tackling a diverse range of problems, we were able to improve out-of-domain performance in all settings. The bias product method works consistently, but can almost always be significantly out-performed by the learned-mixin method with an appropriate entropy penalty. The reweight baseline improved performance on HANS, but was relatively ineffective in other cases. 

Increasing the out-of-domain performance usually comes at the cost of losing some in-domain performance, which is unsurprising since the biased approaches we are removing are helpful on the in-domain data. TriviaQA-CP stands out as a case where this trade-off is minimal.

A possible issue is that our methods reduce the need for the model to solve examples the bias-only model works well on (since the ensemble's prediction will already be mostly correct for those examples), which effectively reduces the amount of training data. An ideal approach would be to block the model from using the bias-only method, and require it to solve examples the bias-only method solves through other means. 
We suspect this will necessitate a more clear-box method since it requires doing fine-grained regularization of how the model is solving individual examples.

\section{Conclusion}
Our key contribution is a method of using human knowledge about what methods will not generalize well to improve model robustness to domain-shift. 
Our approach is to train a robust model in an ensemble with a pre-trained naive model, and then use the robust model alone at test time. Extensive experiments show that our method works well on two adversarial datasets, and two changing-prior datasets, including a 12 point gain on VQA-CP. Future work includes learning to automatically detect dataset bias, which would allow our method to be applicable with less specific prior knowledge.

\section*{Acknowledgements}
This work was supported in part by the ARO (ARO-W911NF-16-1-0121) and the NSF (IIS-1252835, IIS-1562364). We thank Sewon Min, Gabriel Stanovsky, Mandar Joshi and the anonymous reviewers for their helpful comments.

\bibliography{paper.bib}

\begin{thebibliography}{55}
\expandafter\ifx\csname natexlab\endcsname\relax\def\natexlab#1{#1}\fi

\bibitem[{Agrawal et~al.(2018)Agrawal, Batra, Parikh, and Kembhavi}]{vqa_cp}
Aishwarya Agrawal, Dhruv Batra, Devi Parikh, and Aniruddha Kembhavi. 2018.
\newblock {Don't Just Assume; Look and Answer: Overcoming Priors for Visual
  Question Answering}.
\newblock In \emph{CVPR}.

\bibitem[{Anand et~al.(2018)Anand, Belilovsky, Kastner, Larochelle, and
  Courville}]{anand2018blindfold}
Ankesh Anand, Eugene Belilovsky, Kyle Kastner, Hugo Larochelle, and Aaron
  Courville. 2018.
\newblock \href {http://arxiv.org/abs/1811.05013} {{Blindfold Baselines for
  Embodied QA}}.
\newblock \emph{Computing Research Repository}, arXiv:1811.05013.
\newblock Version 1.

\bibitem[{Anderson et~al.(2018)Anderson, He, Buehler, Teney, Johnson, Gould,
  and Zhang}]{bottomuptopdown}
Peter Anderson, Xiaodong He, Chris Buehler, Damien Teney, Mark Johnson, Stephen
  Gould, and Lei Zhang. 2018.
\newblock {Bottom-Up and Top-Down Attention for Image Captioning and Visual
  Question Answering}.
\newblock In \emph{CVPR}.

\bibitem[{Antol et~al.(2015)Antol, Agrawal, Lu, Mitchell, Batra,
  Lawrence~Zitnick, and Parikh}]{vqa1}
Stanislaw Antol, Aishwarya Agrawal, Jiasen Lu, Margaret Mitchell, Dhruv Batra,
  C~Lawrence~Zitnick, and Devi Parikh. 2015.
\newblock {VQA: Visual Question Answering}.
\newblock In \emph{ICCV}.

\bibitem[{Bahdanau et~al.(2015)Bahdanau, Cho, and Bengio}]{bahdanau2014neural}
Dzmitry Bahdanau, Kyunghyun Cho, and Yoshua Bengio. 2015.
\newblock {Neural Machine Translation by Jointly Learning to Align and
  Translate}.
\newblock In \emph{ICLR}.

\bibitem[{Belinkov and Bisk(2018)}]{belinkov2017synthetic}
Yonatan Belinkov and Yonatan Bisk. 2018.
\newblock {Synthetic and Natural Noise Both Break Neural Machine Translation}.
\newblock In \emph{ICLR}.

\bibitem[{Belinkov et~al.(2019)Belinkov, Poliak, Shieber, Van~Durme, and
  Rush}]{belinkov2019adversarial}
Yonatan Belinkov, Adam Poliak, Stuart~M Shieber, Benjamin Van~Durme, and
  Alexander~M Rush. 2019.
\newblock \href {https://doi.org/10.18653/v1/S19-1028} {{On Adversarial Removal
  of Hypothesis-only Bias in Natural Language Inference}}.
\newblock In \emph{StarSem}.

\bibitem[{Bowman et~al.(2015)Bowman, Angeli, Potts, and
  Manning}]{bowman2015large}
Samuel~R Bowman, Gabor Angeli, Christopher Potts, and Christopher~D Manning.
  2015.
\newblock \href {https://doi.org/10.18653/v1/D15-1075} {{A} {L}arge {A}nnotated
  {C}orpus for {L}earning {N}atural {L}anguage {I}nference}.
\newblock In \emph{EMNLP}.

\bibitem[{Burns et~al.(2018)Burns, Hendricks, Saenko, Darrell, and
  Rohrbach}]{burns2018women}
Kaylee Burns, Lisa~Anne Hendricks, Kate Saenko, Trevor Darrell, and Anna
  Rohrbach. 2018.
\newblock {Women Also Snowboard: Overcoming Bias in Captioning Models}.
\newblock In \emph{ECCV}.

\bibitem[{Caglayan et~al.(2019)Caglayan, Madhyastha, Specia, and
  Barrault}]{caglayan2019probing}
Ozan Caglayan, Pranava Madhyastha, Lucia Specia, and Lo{\"\i}c Barrault. 2019.
\newblock \href {https://doi.org/10.18653/v1/N19-1422} {{P}robing the {N}eed
  for {V}isual {C}ontext in {M}ultimodal {M}achine {T}ranslation}.
\newblock In \emph{NAACL}.

\bibitem[{Carlini and Wagner(2018)}]{carlini2018audio}
Nicholas Carlini and David Wagner. 2018.
\newblock {Audio Adversarial Examples: Targeted Attacks on Speech-to-Text}.
\newblock In \emph{2018 {IEEE} Security and Privacy Workshops}.

\bibitem[{Chen and Durrett(2019)}]{chen2019understanding}
Jifan Chen and Greg Durrett. 2019.
\newblock \href {https://doi.org/10.18653/v1/N19-1405} {{Understanding Dataset
  Design Choices for Multi-hop Reasoning}}.
\newblock In \emph{NAACL}.

\bibitem[{Chen et~al.(2017)Chen, Zhu, Ling, Wei, Jiang, and
  Inkpen}]{chen2016enhanced}
Qian Chen, Xiaodan Zhu, Zhenhua Ling, Si~Wei, Hui Jiang, and Diana Inkpen.
  2017.
\newblock \href {https://doi.org/10.18653/v1/P17-1152} {{Enhanced LSTM for
  Natural Language Inference}}.
\newblock In \emph{ACL}.

\bibitem[{Choi et~al.(2018)Choi, He, Iyyer, Yatskar, Yih, Choi, Liang, and
  Zettlemoyer}]{choi2018quac}
Eunsol Choi, He~He, Mohit Iyyer, Mark Yatskar, Wen-tau Yih, Yejin Choi, Percy
  Liang, and Luke Zettlemoyer. 2018.
\newblock \href {https://doi.org/10.18653/v1/D18-1241} {{QuAC: Question
  Answering in Context}}.
\newblock In \emph{EMNLP}.

\bibitem[{Clark and Gardner(2018)}]{clark2017simple}
Christopher Clark and Matt Gardner. 2018.
\newblock \href {https://www.aclweb.org/anthology/P18-1078} {{Simple and
  Effective Multi-Paragraph Reading Comprehension}}.
\newblock In \emph{ACL}.

\bibitem[{Clark et~al.(2018)Clark, Cowhey, Etzioni, Khot, Sabharwal, Schoenick,
  and Tafjord}]{clark2018think}
Peter Clark, Isaac Cowhey, Oren Etzioni, Tushar Khot, Ashish Sabharwal, Carissa
  Schoenick, and Oyvind Tafjord. 2018.
\newblock \href {http://arxiv.org/abs/1803.05457} {{Think you have Solved
  Question Answering? Try ARC, the AI2 Reasoning Challenge}}.
\newblock \emph{Computing Research Repository}, arXiv:1803.05457.
\newblock Version 1.

\bibitem[{Clark et~al.(2016)Clark, Etzioni, Khot, Sabharwal, Tafjord, Turney,
  and Khashabi}]{clark2016combining}
Peter Clark, Oren Etzioni, Tushar Khot, Ashish Sabharwal, Oyvind Tafjord, Peter
  Turney, and Daniel Khashabi. 2016.
\newblock {Combining Retrieval, Statistics, and Inference to Answer Elementary
  Science Questions}.
\newblock In \emph{AAAI}.

\bibitem[{Devlin et~al.(2019)Devlin, Chang, Lee, and
  Toutanova}]{devlin2018bert}
Jacob Devlin, Ming-Wei Chang, Kenton Lee, and Kristina Toutanova. 2019.
\newblock \href {https://doi.org/10.18653/v1/N19-1423} {{BERT: Pre-training of
  Deep Bidirectional Transformers for Language Understanding}}.
\newblock In \emph{NAACL}.

\bibitem[{Edwards and Storkey(2016)}]{edwards2015censoring}
Harrison Edwards and Amos Storkey. 2016.
\newblock {Censoring Representations with an Adversary}.
\newblock In \emph{ICLR}.

\bibitem[{Feng et~al.(2019)Feng, Wallace, and Boyd-Graber}]{feng2019misleading}
Shi Feng, Eric Wallace, and Jordan Boyd-Graber. 2019.
\newblock \href {https://www.aclweb.org/anthology/P19-1554} {{M}isleading
  {F}ailures of {P}artial-input {B}aselines}.
\newblock In \emph{ACL}.

\bibitem[{Finkel et~al.(2005)Finkel, Grenager, and Manning}]{corenlp_ner}
Jenny~Rose Finkel, Trond Grenager, and Christopher Manning. 2005.
\newblock \href {https://doi.org/10.3115/1219840.1219885} {{Incorporating
  Non-local Information into Information Extraction Systems by Gibbs
  Sampling}}.
\newblock In \emph{ACL}.

\bibitem[{Glockner et~al.(2018)Glockner, Shwartz, and
  Goldberg}]{glockner2018breaking}
Max Glockner, Vered Shwartz, and Yoav Goldberg. 2018.
\newblock \href {https://doi.org/10.18653/v1/P18-2103} {{Breaking NLI Systems
  with Sentences that Require Simple Lexical Inferences}}.
\newblock In \emph{ACL}.

\bibitem[{Goyal et~al.(2018)Goyal, Khot, Summers-Stay, Batra, and
  Parikh}]{vqa2}
Yash Goyal, Tejas Khot, Douglas Summers-Stay, Dhruv Batra, and Devi Parikh.
  2018.
\newblock {Making the V in VQA Matter: Elevating the Role of Image
  Understanding in Visual Question Answering}.
\newblock \emph{IJCV}.

\bibitem[{Grand and Belinkov(2019)}]{grand2019adversarial}
Gabriel Grand and Yonatan Belinkov. 2019.
\newblock \href {https://doi.org/10.18653/v1/W19-1801} {{Adversarial
  Regularization for Visual Question Answering: Strengths, Shortcomings, and
  Side Effects}}.
\newblock In \emph{Proceedings of the Second Workshop on Shortcomings in Vision
  and Language}.

\bibitem[{Gururangan et~al.(2018)Gururangan, Swayamdipta, Levy, Schwartz,
  Bowman, and Smith}]{gururangan2018annotation}
Suchin Gururangan, Swabha Swayamdipta, Omer Levy, Roy Schwartz, Samuel~R
  Bowman, and Noah~A Smith. 2018.
\newblock \href {https://www.aclweb.org/anthology/N18-2017} {{Annotation
  Artifacts in Natural Language Inference Data}}.
\newblock In \emph{NAACL}.

\bibitem[{He et~al.(2019)He, Zha, and Wang}]{HeHe}
He~He, Sheng Zha, and Haohan Wang. 2019.
\newblock \href {http://arxiv.org/abs/1908.10763} {{Unlearn Dataset Bias in
  Natural Language Inference by Fitting the Residual}}.
\newblock \emph{Computing Research Repository}, arXiv:1908.10763.
\newblock Version 1.

\bibitem[{Hinton(2002)}]{hinton1999products}
Geoffrey~E. Hinton. 2002.
\newblock {Training Products of Experts by Minimizing Contrastive Divergence}.
\newblock \emph{Neural Computation}.

\bibitem[{Jabri et~al.(2016)Jabri, Joulin, and Van
  Der~Maaten}]{jabri2016revisiting}
Allan Jabri, Armand Joulin, and Laurens Van Der~Maaten. 2016.
\newblock {Revisiting Visual Question Answering Baselines}.
\newblock In \emph{European conference on computer vision}.

\bibitem[{Jia and Liang(2017)}]{adversarial_squad}
Robin Jia and Percy Liang. 2017.
\newblock \href {https://doi.org/10.18653/v1/D17-1215} {{Adversarial Examples
  for Evaluating Reading Comprehension Systems}}.
\newblock In \emph{EMNLP}.

\bibitem[{Joshi et~al.(2017)Joshi, Choi, Weld, and
  Zettlemoyer}]{joshi2017triviaqa}
Mandar Joshi, Eunsol Choi, Daniel~S Weld, and Luke Zettlemoyer. 2017.
\newblock \href {https://doi.org/10.18653/v1/P17-1147} {{TriviaQA: A Large
  Scale Distantly Supervised Challenge Dataset for Reading Comprehension}}.
\newblock In \emph{ACL}.

\bibitem[{Kim et~al.(2019)Kim, Kim, Kim, Kim, and Kim}]{kim2018learning}
Byungju Kim, Hyunwoo Kim, Kyungsu Kim, Sungjin Kim, and Junmo Kim. 2019.
\newblock {Learning Not to Learn: Training Deep Neural Networks with Biased
  Data}.
\newblock In \emph{CVPR}.

\bibitem[{Kingma and Ba(2015)}]{adam}
Diederik~P Kingma and Jimmy Ba. 2015.
\newblock {Adam: A Method for Stochastic Optimization}.
\newblock In \emph{ICLR}.

\bibitem[{Lewis and Fan(2019)}]{generative_qa}
Mike Lewis and Angela Fan. 2019.
\newblock {Generative Question Answering: Learning to Answer the Whole
  Question}.
\newblock In \emph{ICLR}.

\bibitem[{McCoy et~al.(2019)McCoy, Pavlick, and Linzen}]{mccoy2019right}
Tom McCoy, Ellie Pavlick, and Tal Linzen. 2019.
\newblock \href {https://www.aclweb.org/anthology/P19-1334} {{Right for the
  Wrong Reasons: Diagnosing Syntactic Heuristics in Natural Language
  Inference}}.
\newblock In \emph{ACL}.

\bibitem[{Mikolov et~al.(2018)Mikolov, Grave, Bojanowski, Puhrsch, and
  Joulin}]{fasttext_word_vectors}
Tomas Mikolov, Edouard Grave, Piotr Bojanowski, Christian Puhrsch, and Armand
  Joulin. 2018.
\newblock \href {https://www.aclweb.org/anthology/L18-1008} {{A}dvances in
  {P}re-{T}raining {D}istributed {W}ord {R}epresentations}.
\newblock In \emph{LREC}.

\bibitem[{Min et~al.(2019)Min, Wallace, Singh, Gardner, Hajishirzi, and
  Zettlemoyer}]{min2019compositional}
Sewon Min, Eric Wallace, Sameer Singh, Matt Gardner, Hannaneh Hajishirzi, and
  Luke Zettlemoyer. 2019.
\newblock \href {https://www.aclweb.org/anthology/P19-1416} {{Compositional
  Questions Do Not Necessitate Multi-hop Reasoning}}.
\newblock In \emph{ACL}.

\bibitem[{Rajpurkar et~al.(2016)Rajpurkar, Zhang, Lopyrev, and Liang}]{squad}
Pranav Rajpurkar, Jian Zhang, Konstantin Lopyrev, and Percy Liang. 2016.
\newblock \href {https://doi.org/10.18653/v1/D16-1264} {{SQuAD: 100, 000+
  Questions for Machine Comprehension of Text}}.
\newblock In \emph{EMNLP}.

\bibitem[{Ramakrishnan et~al.(2018)Ramakrishnan, Agrawal, and
  Lee}]{ramakrishnan2018overcoming}
Sainandan Ramakrishnan, Aishwarya Agrawal, and Stefan Lee. 2018.
\newblock {Overcoming Language Priors in Visual Question Answering with
  Adversarial Regularization}.
\newblock In \emph{NeurIPS}.

\bibitem[{Reddy et~al.(2019)Reddy, Chen, and Manning}]{reddy2018coqa}
Siva Reddy, Danqi Chen, and Christopher~D. Manning. 2019.
\newblock \href {https://doi.org/10.1162/tacl_a_00266} {{CoQA: A Conversational
  Question Answering Challenge}}.
\newblock In \emph{TACL}.

\bibitem[{Schwartz et~al.(2017)Schwartz, Sap, Konstas, Zilles, Choi, and
  Smith}]{schwartz2017effect}
Roy Schwartz, Maarten Sap, Ioannis Konstas, Li~Zilles, Yejin Choi, and Noah~A
  Smith. 2017.
\newblock \href {https://doi.org/10.18653/v1/K17-1004} {{The Effect of
  Different Writing Tasks on Linguistic Style: A Case Study of the ROC Story
  Cloze Task}}.
\newblock In \emph{CoNLL}.

\bibitem[{Seo et~al.(2017)Seo, Kembhavi, Farhadi, and
  Hajishirzi}]{seo2016bidirectional}
Minjoon Seo, Aniruddha Kembhavi, Ali Farhadi, and Hannaneh Hajishirzi. 2017.
\newblock {BiDirectional Attention Flow for Machine Comprehension}.
\newblock In \emph{ICLR}.

\bibitem[{Smith et~al.(2005)Smith, Cohn, and Osborne}]{smith2005logarithmic}
Andrew Smith, Trevor Cohn, and Miles Osborne. 2005.
\newblock \href {https://doi.org/10.3115/1219840.1219843} {{Logarithmic Opinion
  Pools for Conditional Random Fields}}.
\newblock In \emph{ACL}.

\bibitem[{Srivastava et~al.(2015)Srivastava, Greff, and
  Schmidhuber}]{srivastava2015highway}
Rupesh~Kumar Srivastava, Klaus Greff, and J{\"u}rgen Schmidhuber. 2015.
\newblock Highway networks.
\newblock In \emph{Deep Learning Workshop (ICML)}.

\bibitem[{Suchanek et~al.(2007)Suchanek, Kasneci, and
  Weikum}]{suchanek2007yago}
Fabian~M Suchanek, Gjergji Kasneci, and Gerhard Weikum. 2007.
\newblock {Yago: A Core of Semantic Knowledge}.
\newblock In \emph{WWW}.

\bibitem[{Szegedy et~al.(2014)Szegedy, Zaremba, Sutskever, Bruna, Erhan,
  Goodfellow, and Fergus}]{szegedy2013intriguing}
Christian Szegedy, Wojciech Zaremba, Ilya Sutskever, Joan Bruna, Dumitru Erhan,
  Ian Goodfellow, and Rob Fergus. 2014.
\newblock {Intriguing Properties of Neural Networks}.
\newblock In \emph{ICLR}.

\bibitem[{Wang and Jiang(2017)}]{wang2016machine}
Shuohang Wang and Jing Jiang. 2017.
\newblock {Machine Comprehension Using Match-LSTM and Answer Pointer}.
\newblock In \emph{ICLR}.

\bibitem[{Wang et~al.(2018)Wang, Zhao, Chang, Yatskar, and
  Ordonez}]{wang2018adversarial}
Tianlu Wang, Jieyu Zhao, Kai-Wei Chang, Mark Yatskar, and Vicente Ordonez.
  2018.
\newblock \href {http://arxiv.org/abs/1811.08489} {{Adversarial Removal of
  Gender from Deep Image Representations}}.
\newblock \emph{Computing Research Repository}, arXiv:1811.08489.
\newblock Version 2.

\bibitem[{Wang and Bansal(2018)}]{wang2018robust}
Yicheng Wang and Mohit Bansal. 2018.
\newblock \href {https://doi.org/10.18653/v1/N18-2091} {{Robust Machine
  Comprehension Models via Adversarial Training}}.
\newblock In \emph{NAACL}.

\bibitem[{Welbl et~al.(2018)Welbl, Stenetorp, and
  Riedel}]{welbl2018constructing}
Johannes Welbl, Pontus Stenetorp, and Sebastian Riedel. 2018.
\newblock \href {https://doi.org/10.1162/tacl_a_00021} {{Constructing Datasets
  for Multi-hop Reading Comprehension across Documents}}.
\newblock \emph{TACL}.

\bibitem[{Yang et~al.(2018)Yang, Qi, Zhang, Bengio, Cohen, Salakhutdinov, and
  Manning}]{hotpotqa}
Zhilin Yang, Peng Qi, Saizheng Zhang, Yoshua Bengio, William~W Cohen, Ruslan
  Salakhutdinov, and Christopher~D Manning. 2018.
\newblock \href {https://doi.org/10.18653/v1/D18-1259} {{H}otpotqa: {A}
  {D}ataset for {D}iverse, {E}xplainable {M}ulti-hop {Q}uestion {A}nswering}.
\newblock In \emph{EMNLP}.

\bibitem[{Zellers et~al.(2018)Zellers, Bisk, Schwartz, and Choi}]{swag}
Rowan Zellers, Yonatan Bisk, Roy Schwartz, and Yejin Choi. 2018.
\newblock \href {https://doi.org/10.18653/v1/D18-1009} {{S}wag: {A}
  {L}arge-{S}cale {A}dversarial {D}ataset for {G}rounded {C}ommonsense
  {I}nference}.
\newblock In \emph{EMNLP}.

\bibitem[{Zhang et~al.(2018{\natexlab{a}})Zhang, Lemoine, and
  Mitchell}]{zhang2018mitigating}
Brian~Hu Zhang, Blake Lemoine, and Margaret Mitchell. 2018{\natexlab{a}}.
\newblock {Mitigating Unwanted Biases with Adversarial Learning}.
\newblock In \emph{AIES}.

\bibitem[{Zhang et~al.(2016)Zhang, Goyal, Summers-Stay, Batra, and
  Parikh}]{zhang2016yin}
Peng Zhang, Yash Goyal, Douglas Summers-Stay, Dhruv Batra, and Devi Parikh.
  2016.
\newblock {Yin and Yang: Balancing and Answering Binary Visual Questions}.
\newblock In \emph{CVPR}.

\bibitem[{Zhang et~al.(2018{\natexlab{b}})Zhang, Liu, Liu, Gao, Duh, and
  Van~Durme}]{record}
Sheng Zhang, Xiaodong Liu, Jingjing Liu, Jianfeng Gao, Kevin Duh, and Benjamin
  Van~Durme. 2018{\natexlab{b}}.
\newblock \href {https://arxiv.org/abs/1810.12885} {{R}e{C}o{R}{D}: {B}ridging
  the {G}ap between {H}uman and {M}achine {C}ommonsense {R}eading
  {C}omprehension}.
\newblock \emph{Computing Research Repository}, arXiv:1810.12885.
\newblock Version 1.

\bibitem[{Zhao et~al.(2017)Zhao, Wang, Yatskar, Ordonez, and
  Chang}]{zhao2017men}
Jieyu Zhao, Tianlu Wang, Mark Yatskar, Vicente Ordonez, and Kai-Wei Chang.
  2017.
\newblock \href {https://doi.org/10.18653/v1/D17-1323} {{Men Also Like
  Shopping: Reducing Gender Bias Amplification using Corpus-level
  Constraints}}.
\newblock In \emph{EMNLP}.

\end{thebibliography}
\bibliographystyle{acl_natbib}

\appendix
\section{Entropy Penalty Weights}
\label{appendix:penalty_weights}
The strength of the entropy penalty used for the learned-mixin +H model can be found in Table~\ref{tab:hyperparameters}.

\begin{table}[]
    \centering
    \begin{tabular}{l|l|c}
        Dataset & Experiment & Penalty \\ \hline
        Synthetic & Indicator & 0.01 \\
        Synthetic & Excluder & 0.005 \\
        Synthetic & Dependent & 0.005 \\
        VQA-CP & - & 0.36 \\
        HANS & Recurrent & 0.03 \\
        HANS & BERT & 0.03 \\
        Adver. SQuAD & TF-IDF Filtered & 2.0 \\
        Adver. SQuAD & TF-IDF & 2.0 \\
        TriviaQA-CP & Location & 0.4 \\
        TriviaQA-CP & Person & 0.2 \\
    \end{tabular}
    \caption{Entropy penalty weight for the learned-mixin +H ensemble on all our experiments.}
    \label{tab:hyperparameters}
\end{table}

\section{Co-Attention NLI Model}
\label{appendix:nli_model}
\label{sect:co-attention-model}
The model we use for NLI is based on ESIM~\cite{chen2016enhanced}. It has the following stages:
\\
\\
\noindent
\textbf{Embed}: Embed the words using a character CNN, following what was done by \citet{seo2016bidirectional}, and the fasttext crawl word embeddings~\cite{fasttext_word_vectors}, then run a shared BiLSTM over the results.
\\
\\
\noindent
\textbf{Co-Attention}: Compute an attention matrix using the formulation from~\citet{seo2016bidirectional}, and use it to compute a context vector for each premise word~\cite{bahdanau2014neural}. Then build an augmented vector for each premise word by concatenating the word's embedding, the context vector, and the elementwise product of the two.
Augmented vectors for the hypothesis are built in the same way using the transpose of the attention matrix.
\\
\\
\noindent
\textbf{Pool}: Run another shared BiLSTM over the augmented vectors, and max-pool the results. 
The max-pooled vectors from the premise and hypothesis are fed into a fully-connected layer, and then into a softmax layer with three outputs to compute class probabilities.
\\
\\
We apply variational dropout at a rate of 0.2 between all layers, and to the recurrent states of the LSTM, and train the model for 30 epochs using the Adam optimizer~\cite{adam} with a batch size of 32. The learning rate is decayed by 0.999 every 100 steps. We use 200 dimensional LSTMs and a 50 dimensional fully connected layer.

\section{Fine-Grained HANS Results}
We show the scores our methods achieve for the various heuristics used in HANS in Table~\ref{tab:fine-grained-hans}. Our methods reduce the extent to which models naively guess entailment in all cases. Interestingly, the BERT model shows significantly degraded performance on the entailment examples when using the reweight and bias product method, but largely maintains its performance on those examples when using the learned-mixin method.

\label{sect:fine-grained-hans}
\label{appendix:fine_grained_hans}
\begin{table*}[]
    \centering
    \tablefont
\begin{tabular}{llccccccc} \toprule
\multirow{2}{*}{Model} & \multirow{2}{*}{Debiasing Method} & \multirow{2}{*}{MNLI} & \multicolumn{3}{c}{Correct: \textit{Entailment}} & \multicolumn{3}{c}{Correct: \textit{Non-entailment}}\\
\cmidrule(r){4-6} \cmidrule(l){7-9}
 &  &  & Lexical & Subseq. & Const & Lexical & Subseq. & Const\\ \midrule
 \multirow{5}{*}{Co-Attention} & None & 78.73 & 97.83 & 99.67 & 97.28 & 1.37 & 3.68 & 3.68 \\
 & Reweight & 77.03 & 80.10 & 77.84 & 73.76 & 15.68 & 34.27 & 35.44 \\
 & Bias Product & 76.63 & 77.89 & 76.61 & 70.95 & 17.89 & 35.11 & 43.71 \\
 & Learned-Mixin & 78.05 & 94.84 & 97.25 & 91.19 & 3.69 & 9.57 & 13.37 \\
 & Learned-Mixin +H & 74.50 & 67.18 & 61.05 & 47.13 & 27.41 & 56.82 & 60.53 \\
\hdashline
\multirow{5}{*}{BERT} & None & 84.24 & 96.30 & 99.58 & 99.30 & 49.03 & 7.88 & 22.30 \\
 & Reweight & 83.54 & 67.93 & 84.34 & 80.97 & 77.44 & 44.87 & 59.57 \\
 & Bias Product & 82.97 & 53.67 & 69.47 & 70.88 & 81.34 & 62.93 & 69.23 \\
 & Learned-Mixin & 84.29 & 95.64 & 99.52 & 99.14 & 55.41 & 8.34 & 25.96 \\
 & Learned-Mixin +H & 83.97 & 91.98 & 98.20 & 97.98 & 64.99 & 13.25 & 30.48 \\
\bottomrule
\end{tabular}
    \caption{Scores on individual heuristics in HANS, with scores on the MNLI matched dev set for reference. Results are an average of 8 runs.}
    \label{tab:fine-grained-hans}
    
\end{table*}

\section{Modified BiDAF QA Model}
\label{appendix:bidaf-model}
The model we use for QA is based on BiDAF~\cite{seo2016bidirectional}. It has the following stages:
\\
\\
\noindent
\textbf{Embed}: Embed the words using a character CNN following~\citet{seo2016bidirectional} and the fasttext crawl word embeddings~\cite{fasttext_word_vectors}. Then run a BiLSTM over the results to get context-aware question embeddings and passage embeddings.
\\
\\
\noindent
\textbf{Bi-Attention}: Apply the bi-directional attention mechanism from~\citet{seo2016bidirectional} to produce question-aware passage embeddings.
\\
\\
\noindent
\textbf{Predict}: Apply a fully connected layer, then two more BiLSTM layers, then a two dimensional linear layer to produce start and end scores for each token.
\\
\\
We apply variational dropout at a rate of 0.2 between all layers. We use the Adam optimizer~\cite{adam} with a batch size of 45, while decaying the learning rate by 0.999 every 100 steps.

For SQuAD, we use a 200 dimensional fully connected layer and 100 dimensional LSTMs.

For TriviaQA we use a 256 dimensional fully connected layer and 128 dimensional LSTMs, with highway connections between each BiLSTM~\cite{srivastava2015highway} and a recurrent dropout rate of 0.2.

\section{TriviaQA-CP}
\label{appendix:triviaqa_cp}
\begin{table}[]
    \centering
    \tablefont
\begin{tabular}{lcc} \toprule
Statistic & Location & Person\\ \midrule
Num Train & 60,133 & 52,953 \\ 
Num Test & 1,992 & 2,865 \\ 
Avg. Passage Length & 318 & 317 \\ 
Avg. Question Length & 16.7 & 16.0 \\ 
\bottomrule
\end{tabular}
    \caption{Statistics for the TriviaQA-CP datasets.}
    \label{tab:triviqa-stats}
\end{table}

\begin{table*}
    \centering
    \tablefont
\begin{tabular}{lcccccccc} \toprule
\multirow{2}{*}{Method} & \multirow{2}{*}{Accuracy} & \multicolumn{3}{c}{Location} & \multicolumn{3}{c}{Person}\\ 
\cmidrule(r){3-5} \cmidrule(l){6-8}
 & & Precision & Recall & F1 & Precision & Recall & F1\\ \midrule
Patterns & 72.84 & 98.30 & 70.61 & 82.19 & 99.12 & 33.43 & 50.00 \\ 
Yago & 88.56 & 95.87 & 85.31 & 90.28 & 94.70 & 80.00 & 86.73 \\ 
Yago + Patterns & 91.73 & 95.44 & 93.88 & 94.65 & 94.70 & 85.37 & 89.80 \\ 
Dist. Supervised Model & 92.73 & 96.60 & 92.65 & 94.58 & 98.28 & 85.37 & 91.37 \\ 
Supervised Model & 94.46 & 95.08 & 94.69 & 94.89 & 93.31 & 95.82 & 94.55 \\ 
\bottomrule
\end{tabular}
    \caption{Accuracy, and per-class scores, on the manually annotated questions for the various question classification methods we used when building TriviaQA-CP.}
    \label{tab:question-clf-results}
\end{table*}

In this section we discuss our changing-prior TriviaQA dataset, TriviaQA-CP. This dataset was built by training a classifier to identify TriviaQA~\cite{joshi2017triviaqa} questions as being about people, locations, or other topics, and then selecting an answer-containing passage for each question as context. There are two versions of this dataset: a person changing-priors dataset that was built by removing the person questions from the train set and using only person questions from the dev set for evaluation, and a location changing-priors dataset that was built by repeating this process for location questions. Statistics for these two sets are shown in Table~\ref{tab:triviqa-stats}. We review the three-step procedure we used to construct this dataset below.
\\
\\ 
\noindent \textbf{Distantly Supervised Classification}: We first train a preliminary question-type classifier using distant supervision. We noisily label person and location questions using a manually constructed set of patterns (e.g., questions with the phrase ``What is the family name of..." are almost always about people), and by attempting to look up the answers in the Yago database~\cite{suchanek2007yago} and checking if the answer belongs to a person or location category. Questions that did not match either of these heuristics are labelled as other.

We use these labels to train a simple recurrent model that embeds the question using the fasttext words vectors, applies a 100 dimensional BiLSTM, max-pools, and then applies a softmax layer with 3 outputs. We train the model for 3 epochs using the Adam optimizer~\cite{adam}, and apply 0.5 dropout to the embeddings and 0.2 dropout to the recurrent states and the output of the max-pooling layer.
\\
\\
\noindent \textbf{Supervised Classification}: Next we use higher quality labels to train a second linear classifier to re-calibrate the recurrent model's predictions, and to integrate its predictions with the distantly supervised heuristics. 
An author manually labelled 1,100 questions, then a classifier was trained on those questions using the predictions from the recurrent model as features, as well as two additional features built from looking up the category of the answer in Yago as before. This classifier was then used to decide the final question classifications.

Table~\ref{tab:question-clf-results} shows the accuracy of these classifiers. The final model achieves about 95\% accuracy. We find about 25\% of the questions are about people and about 20\% of the questions are about locations.
\\
\\
\noindent \textbf{Paragraph Selection}: In TriviaQA, each question is paired with multiple documents. We simplify the task by selecting a single answer-containing paragraph for each question. We use the approach of~\citet{clark2017simple} to break up the documents into passages of at most 400 tokens, and rank the passages for each question using their linear paragraph ranker. Each question is then paired with the highest ranking paragraph that contains an answer.

\label{sect:triviaqa-cp}

\end{document}